\documentclass[letterpaper, 10 pt, conference,final]{ieeeconf} 

\IEEEoverridecommandlockouts                              
\usepackage[english]{babel}
\usepackage[utf8]{inputenc}
\usepackage{graphicx} 

\usepackage{amsmath}
\usepackage{amsfonts}
\usepackage[dvipsnames]{xcolor}
\usepackage{multirow}
\usepackage[]{ragged2e}
\usepackage{hyperref}
\usepackage{textcomp}
\usepackage{tikz}

\DeclareMathOperator*{\argmax}{arg\,max}
\newtheorem{theorem}{Theorem}

\usetikzlibrary{shapes.geometric}
\usetikzlibrary{calc}
\newcommand{\IEEEcopyright}{
   \begin{tikzpicture}[remember picture,overlay]
     \node[align=left, anchor=north west]
       at ($(current page.north west) + (1,-1)$)
       { \textcopyright 2019 IEEE CASE.  Personal use of this material is permitted.  Permission from IEEE must be obtained for all other uses, in any current \\ or future media, including reprinting/republishing this material for advertising or promotional purposes, creating new collective \\works, for resale or redistribution to servers or lists, or reuse of any copyrighted component of this work in other works.};
   \end{tikzpicture}
}

\title{\LARGE \bf
Mean Spectral Normalization of Deep Neural Networks for Embedded Automation
}

\author{Anand Krishnamoorthy Subramanian$^{1}$ and Nak Young Chong$^{1}$
\thanks{$^{1}$The authors are with the School of Information Science, Japan Advanced Institute of Science and Technology, Ishikawa, Japan
        {\tt\small \{anandkrish, nakyoung\}@jaist.ac.jp}}%
}

\begin{document}

\maketitle

\IEEEcopyright{}
\thispagestyle{empty}
\pagestyle{empty}

\begin{abstract}
Deep Neural Networks (DNNs) have begun to thrive in the field of automation systems, owing to the recent advancements in standardising various aspects such as architecture, optimization techniques, and regularization. In this paper, we take a step towards a better understanding of \textit{Spectral Normalization} (SN) and its potential for standardizing  regularization of a wider range of Deep Learning models, following an empirical approach. We conduct several experiments to study their training dynamics, in comparison with the ubiquitous \textit{Batch Normalization} (BN) and show that SN increases the gradient sparsity and controls the gradient variance. Furthermore, we show that SN suffers from a phenomenon, we call the \textit{mean-drift} effect, which mitigates its performance. We, then, propose a weight reparameterization called as the \textit{Mean Spectral Normalization} (MSN) to resolve the mean drift, thereby significantly improving the network's performance. Our model performs $\sim 16\%$ faster as compared to BN in practice, and has fewer trainable parameters. We also show the performance of our MSN for small, medium, and large CNNs $-$ 3-layer CNN, VGG7 and DenseNet-BC, respectively $-$ and unsupervised image generation tasks using Generative Adversarial Networks (GANs)  to evaluate its applicability for a broad range of embedded automation tasks.
\end{abstract}

\section{Introduction}
The rise of application of Deep Neural Networks (DNNs) to robot automation motivates various research questions that typically differ from that of the traditional computer vision. While the research direction in DNNs mainly focus towards building architectures, developing loss functions and studying their internal mechanism, standardizing DNN models has been the main motivation and an essential criterion for their successful application to robot automation. The reason being that traditional robot automation relies on well-understood white-box models, while DNNs are black-box models where progress is still being made to reach a consensus about their internal behaviour and dynamics. By \textit{standardizing}, we mean the basic well-behaved framework of architectures, loss functions, regularization techniques and activation functions for the DNN models.

To reinforce our motivation, we shall provide some recent examples of applications of DNNs in automation tasks. Concerning industrial automation, DNNs have found applications that include fault diagnosis \cite{chadha2017comparison}, combustion optimization \cite{cheng2018deep}, welding faults detection \cite{choi2018localization}, traffic control \cite{fadlullah2017state}, power line inspection \cite{nguyen2018automatic} and spectrum sensing \cite{davaslioglu2018generative}. Even Generative Adversarial Networks (GANs) have attracted some attention and have been applied in practice (apart from the standard unsupervised image generation or translation tasks), such as fault detection \cite{chakraborty2018generative}. The main surge in the interests and applications of DNNs can be reasoned as follows - unlike traditional/classical automation tasks where the \textit{features} or the control elements of the task are usually preset and their dynamics/state changes are completely characterised analytically, DNNs try to capture the relevant features and their dynamics automatically given the nominally pre-processed data. The characteristics of DNNs where this feature selection and extraction happens due to their hierarchical structure and their ability to train over modern hardware over thousands of data points are the foremost attractive reasons for their current popularity and success. 

The above discussed automation systems often use small-medium sized neural networks in their tasks owing to speed and hardware/cost constraints. DNNs are over-parameterised models, in the sense that they have large number of trainable parameters (typically hundreds of thousands) compared to the size of the dataset ($\sim$ tens of thousands). Though this over-parameterization helps greatly in optimizing the network weights \cite{li2018learning}, regularization methods are required to improve the generalization and stability of the network during training. As such, a lot of regularization methods have been developed to address this problem; of which the class of methods called \textit{Weight reparameterization} have proven to be quite successful.

Weight reparameterization techniques like Batch Normalization (BN) \cite{ioffe2015batch}, Weight Normalization (WN) \cite{salimans2016weight}, Layer Normalization (LN) \cite{ba2016layer} are implicit regularization methods that restrict the capacity of the over-parameterized network by normalizing/reparameterizing the network weights. Among them, BN has established itself as a very effective component of almost all modern DNNs. This is backed by its stability over a wide range of learning rates, ability to train over large minibatch sizes and faster convergence.

Although BN works well for almost all architectures of neural networks, it is an overkill for small-medium sized networks as it introduces additional training parameters. It is in this background that we investigate the application of spectral normalization (SN) for such networks $-$ owing to it requiring no additional parameters; unlike batch normalization. SN \cite{miyato2018spectral} is a recently introduced technique for normalizing the Lipschitz constant of intermediate layers of deep neural networks, originally proposed for Wasserstein GANs. We, however, have found that SN  performs poorly compared to BN for small to medium-sized networks, and identified the cause of which to be the effect called \textit{mean-drift}. We rectify this effect using our proposed \textit{Mean Spectral Normalization} (MSN), thereby improving the performance to be  comparable to that of BN.

We demonstrate empirically that our MSN method works across a wide range of model depths with fewer parameters and performs at par with BN. Most of the recent applications of neural networks to automation tasks utilise smaller to medium neural networks with 3-20 layers - usually convolution or fully connected layers \cite{zhang2019bearing}. Almost every neural network model employs BN (or its slight variants) for regularization and training stabilization. We, therefore, extend our ideas to a wide range of models, even to very deep networks, with greatly improved performance. Through our method, we propose to standardize the regularization aspect of DNNs, and applicable for robot automation tasks. Our contributions can be summarized as follows:

\begin{figure*}[ht]
    \centering
    \includegraphics[width = 0.8\textwidth, height=5.0cm]{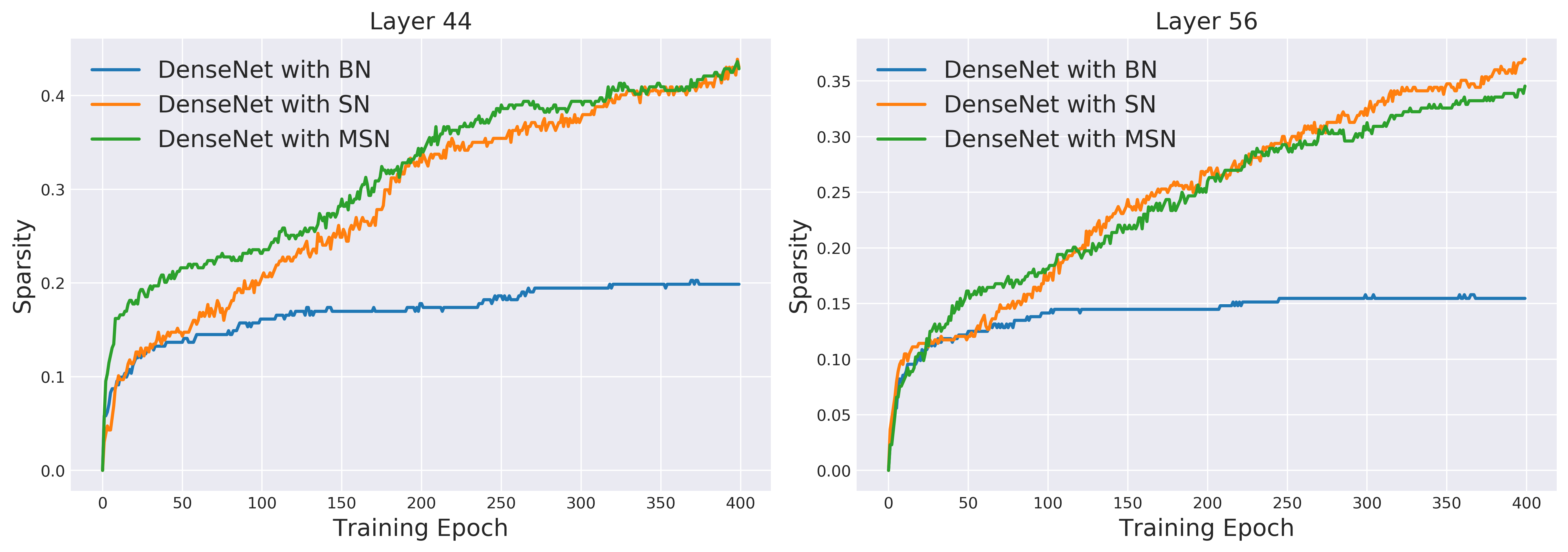}
    \caption{Sparsity of gradients during training DenseNet-BC with LeakyReLU activation function. SN immediately induces a high percent of sparsity to the gradients of the model and steadily increases the sparsity of the gradients during training.}
    \label{fig:sparse_gradients_cifar10}
\end{figure*} 

\begin{itemize}
    \itemsep0em
    \item We provide empirical results for the sparsity of gradients in spectral normalized networks (Refer to Fig. \ref{fig:sparse_gradients_cifar10}).  By bounding the gradient magnitude of the activations (Lipschitz normalization), spectral normalized networks yield a much sparser network compared to the sparsity induced by the rectified linear units (ReLU).
    \item We show that by controlling the mean layer singular values, spectral norm offers better utilization of feature dimensions, unlike other methods such as weight normalization.
    \item We identify the \textit{mean-drift} effect to be a major cause for the diminishing performance of SN as a regularization technique for small and medium sized networks.
    \item We propose a modified SN technique called \textit{Mean Spectral Normalization}(MSN) to correct for the mean-drift and accelerate the performance of the spectral normalized network for small, medium and large neural networks.
\end{itemize}

The structure of this paper is as follows - Section \ref{sec:backgrnd} introduces BN and SN methods formally; In  Section \ref{sec:msn}, we introduce our mean spectral normalization. In Section \ref{sec:exp}, we provide our experimental results (focussing on image-related tasks) and various empirical observations. Note that in this paper, we follow the current trend of empirical insights into deep learning to provide solid experimental footing to understand the dynamics of the SN.

\section{Background} \label{sec:backgrnd}
In this section we discuss batch normalization and spectral normalization techniques for convolutional neural networks along with some background about other weight reparameterization methods. The normalization methods discussed here, come under a subclass of regularisation methods where the network parameters are normalised based on some norm of their parameters which limit some of their capabilities. For example, the $L_2$ norm tends to limit the parameters values to lie on a unit ball centered about the origin {\it i.e.} be closer to zero. The key insight here is that a normalization method makes the network invariant to the scaling of the weights. This makes the network more robust to the new data points and parameter initialization strategies. This is true for all currently used normalization techniques as well as our proposed one.

\subsection{Batch Normalization}
Batch Normalization (BN) or simply batch norm, was initially introduced to reduce the \textit{internal covariate-shift} (ICS) in DNNs. The internal covariate shift is the phenomenon when the distribution of activations $\mathbf{g}_k$ of a layer $k$ shift due to the weight updates in the previous layers during training. Batch norm rectifies this problem by simply standardizing(z-score normalization) the activations of the intermediate layers to zero mean and unit variance and rescaling them using the affine transformations $\gamma_k$ and $\beta_k$ along each channel with respect to all the pixels/points in the input tensor.
\begin{align}\label{eq:BN}
    \mathbf{z}_k = \gamma_k \frac{\mathbf{g}_k - \mathbb{E}_c [\mathbf{g}_k]}{\sqrt{Var(\mathbf{g}_k) + \epsilon}} + \beta_k
\end{align}
where $\epsilon$ is a small value adding for numerical stability. By standardizing the layer weights, we essentially remove the dependencies on the previous layer updates. Rescaling the weights based on some learnable parameters ($\gamma_k$, $\beta_k$) -called the scaling factor and bias respectively- enable the flexibility in choosing appropriate weights during the training. Note that the batch normalized activations have a norm independent of the data, and depends only on the effective layer dimension $D$ and the affine scaling $\gamma_k$. The effective layer dimension $D$ for a input tensor of dimensions $m \times H \times W \times C$ is simply $D:= \sqrt{mHWC}$ where $m$ is the minibatch size, $H, W,C$ are the height, width and depth (number of channels) of the input data.

The concrete reasoning behind the exceptional performance of BN is still being investigated. The recent findings include preventing gradient explosion, improving optimization by smoothing the loss landscape \cite{zhang2016understanding} and, most importantly, improving the Lipschitzness of the layer \cite{santurkar2018does}, {\it i.e.}, the gradients become more concentrated around the mean. This reduction of gradient variance has been accepted \cite{bjorck2018understanding} as  one of the core reasons for the success of BN. The idea of controlling the Lipschitzness of the network, motivated us to probe a related weight reparameterization technique - SN. Furthermore, in practice, BN is difficult to accelerate as it is bounded by memory-bandwidth. Precisely, BN requires two passes through the input data to compute the statistics of the minibatch and then to normalize the output; and this may consume up to a quarter of the total training time for large networks \cite{gitman2017comparison}. Other similar normalization methods like Layer normalization and Instance normalization \cite{ulyanov2016instance} are slight variations to BN with normalization across different dimensions of the output like channels, layers or spatial dimensions. As such, all these suffer from the same drawbacks as that of BN.

\subsection{Spectral Normalization}
Spectral Normalization (SN) \cite{miyato2018spectral} essentially restricts Lipschitz constant of the network to unity by restricting the spectral norm of each layer. Recall that the a function $f$ is $K-$Lipschitz if $|f(x_1) - f(x_2)| \leq K \|x_1 - x_2\|$, for all $x_1, x_2$; where $K$ is the Lipschitz norm of $f$. In other words, small changes in the input of the function causes corresponding small changes in the magnitude of its gradients. From its definition, the Lipschitz constant of a given intermediate layer $k$ of a neural network, whose activations are given by $\mathbf{g}_k = f( \mathbf{W}\mathbf{g}_{k-1})$, is equal to the spectral norm $ \sigma(\mathbf{W})$ of the weight matrix $\mathbf{W}$. Here, $f$ is the activation function. The SN method is then, defined as follows.
\begin{align}\label{eq:SN}
    \hat{\mathbf{W}} = \mathbf{W}/ \sigma(\mathbf{W}),
\end{align}
where $\sigma(\mathbf{W})$ is the spectral norm ($L_2$ matrix norm) of the weight matrix $\mathbf{W}$ given by
\begin{align}
    \sigma(\mathbf{W}) = \sup_{\mathbf{g}_{k-1}} \sigma (\nabla \mathbf{g}_k) =  \max_{\mathbf{x} \neq 0} \frac{\|\mathbf{W} \mathbf{x}\|_2}{\|\mathbf{x}\|_2} 
\end{align}
Again, from its definition, the spectral norm is essentially the largest singular value of the matrix $\mathbf{W}$. Furthermore, the spectral normalization of each layer applies to the weights of the layer and not the activation; similar to the Weight Normalization techniques. This is a crucial distinction from BN which applies to the activations. Since the weights are much fewer than the activation of intermediate layers, SN is often computationally faster than BN. In practice, this means that SN is not bounded by memory-bandwidth, unlike BN. An important caveat is that the Lipschitz norm of the activation function used must be equal to $1$. Therefore, we are limited to activations such as ReLU and leaky ReLU \footnote{For proof, refer to Lemma A.1 in the Appendix of \cite{bartlett2017spectrally}}.

In contrast to spectral norm regularization \cite{yoshida2017spectral}, which penalizes the spectral norm by adding an explicit regularization term to the loss function, the layer weights are simply divided by their corresponding spectral norm in SN. Furthermore, convolutional neural networks usually have fewer weights compared to pre-activations. Therefore, SN is computationally much cheaper, and does not introduce any additional parameters to be trained as evident from Eq. (\ref{eq:SN}).

\begin{figure*}[ht]
    \centering
    \includegraphics[width = 0.9\textwidth, height=5cm]{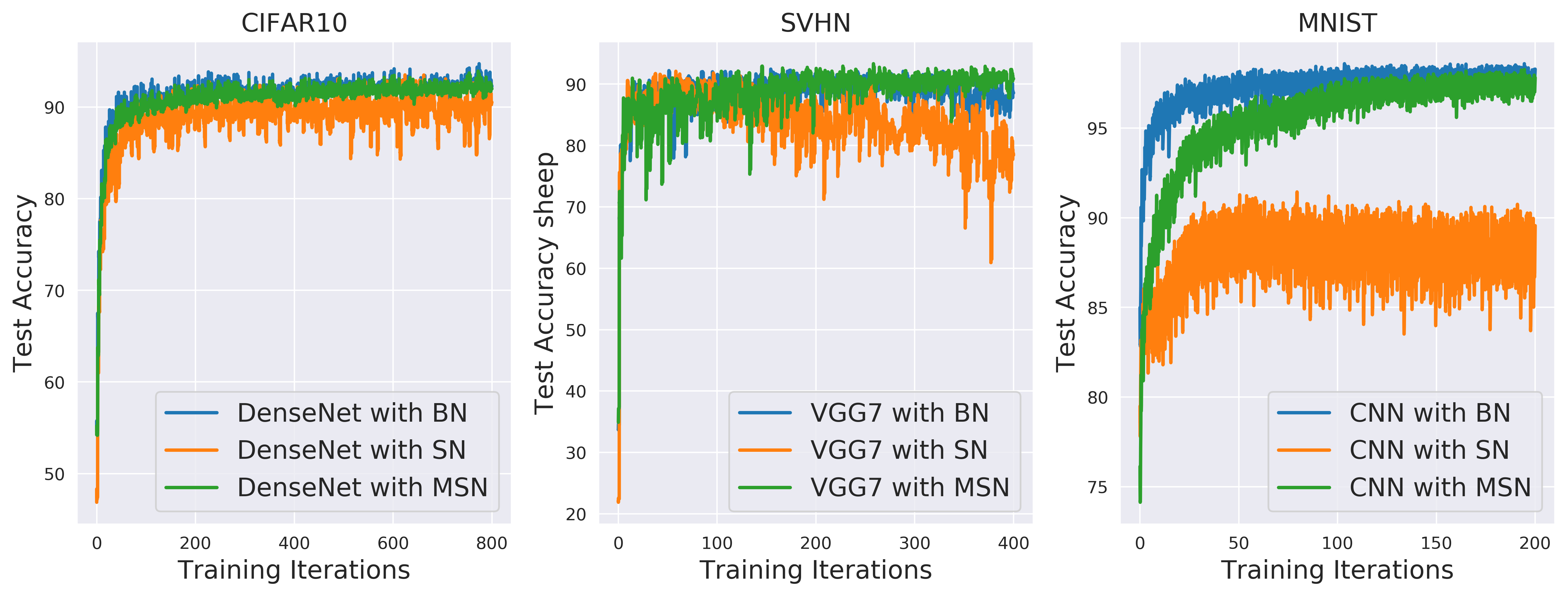}
    \caption{Test Accuracy of various normalization methods during training DenseNet-BC, 3-layer CNN and VGG-7 models (with learning rate 0.001). We observe that SN perform poorly for shallow CNNs, though the performance improves with the depth of the neural network (Compare VGG-7 and DenseNet-BC). MSN clearly improves upon SN for both shallow and deep neural networks and performs comparable to BN.}
    \label{fig:test_acc_cifar10}
\end{figure*}

\begin{figure*}[ht]
    \centering
    \includegraphics[width = 0.9\textwidth, height=5cm]{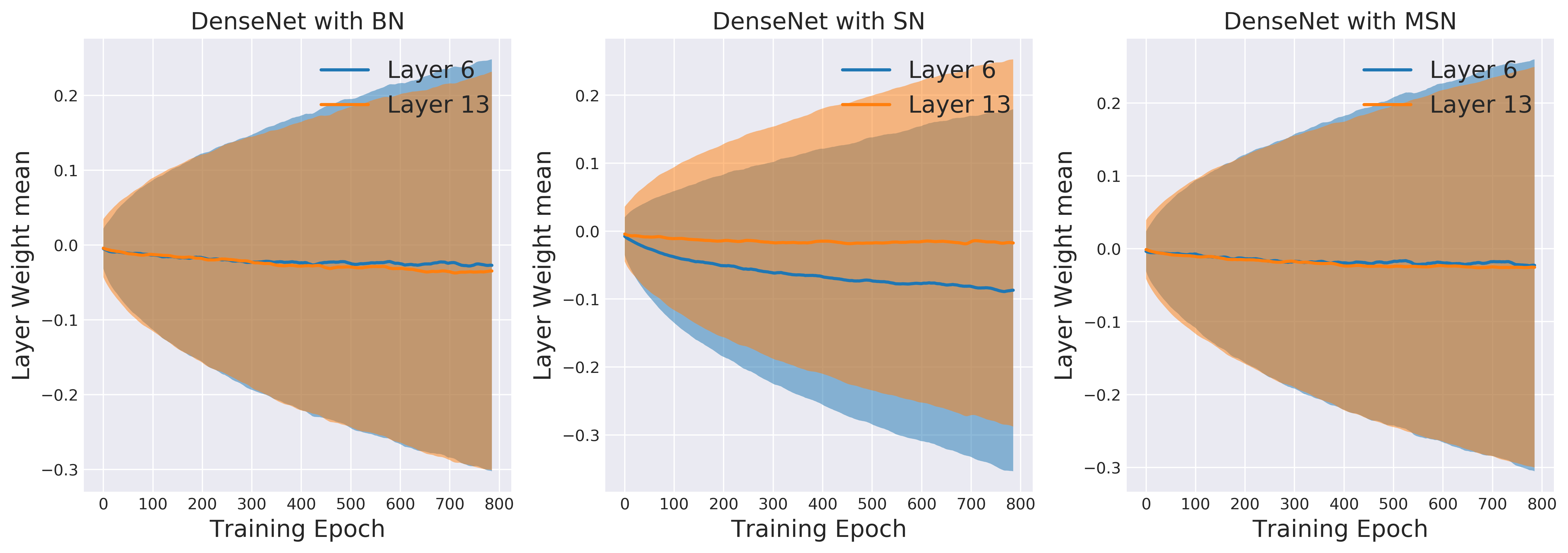}
    \caption{Summary statistics of layer weights showing the internal covariate shift for various normalization methods while training DenseNet-BC. During training, BN controls the mean and variance of the weight distribution while SN only controls the variance. Our proposed MSN corrects for the drift in the mean and variance, thereby improving the performance of the network.}
    \label{fig:layer_mean_cifar10}
\end{figure*}

\section{Proposed Method}\label{sec:msn}
\subsection{Motivation}
In this section, we provide some stronger theoretical motivation for the use of spectral norm for regularizing DNNs. Recent theoretical insights \cite{bartlett2017spectrally}, \cite{neyshabur2017exploring} in analyzing the learning capacity and the generalizability of the neural networks have shown that the those characteristics can be bounded by the network's \textit{spectral complexity} $\mathcal{R}$. The spectral complexity of a given neural network is given by
\begin{align}
    \mathcal{R} := \bigg (\prod_{k=1}^L \sigma(\mathbf{W}_k)\bigg ) \bigg (\sum_{k=1}^L \phi(\mathbf{W}_k) \bigg )^{3/2}
\end{align}
which is essentially the product of the spectral norm of the weight matrices $\mathbf{W}_k$ of all the $L$ layers in the network times a correction factor $\phi$ dependent on those weight matrices.

\begin{theorem}\label{th:sn_bound}
For a given neural network function $F$, computed as $F(\mathbf{x}) = f_L(\mathbf{W}_L f_{L-1}(\mathbf{W}_{L-1}\cdots f_1(\mathbf{W}_1\mathbf{x})\cdots))$ where $f_k$ is the activation function at layer $k$ and a dataset $\{\mathbf{x}_i, y_i\}_{i=1}^N$, drawn i.i.d from some data distribution, we have
\begin{align}
    P[\argmax_{i} F(\mathbf{x})_i \neq y] \leq \mathcal{O} \big ( \mathcal{R}\big ) + \mathcal{E}(F)
\end{align}
where $\mathcal{E}(F)$ is the risk associated with the network, defined as the expectation of the loss.
\end{theorem}
The above theorem states that the generalization error can be reduced by reducing the upper bound, given by the spectral complexity of the network\footnote{For a complete proof of Theorem \ref{th:sn_bound}, refer to \cite{bartlett2017spectrally}}. Additionally, recent studies \cite{yoshida2017spectral}, \cite{gouk2018regularisation} have been conducted in enforcing such Lipschitz continuity in neural networks, albeit through explicit regularization methods, in contrast to our proposed implicit technique.

\subsection{Mean Drift}
\begin{figure}[ht]
    \centering
    \includegraphics[width = 0.9\textwidth, height=5.0cm]{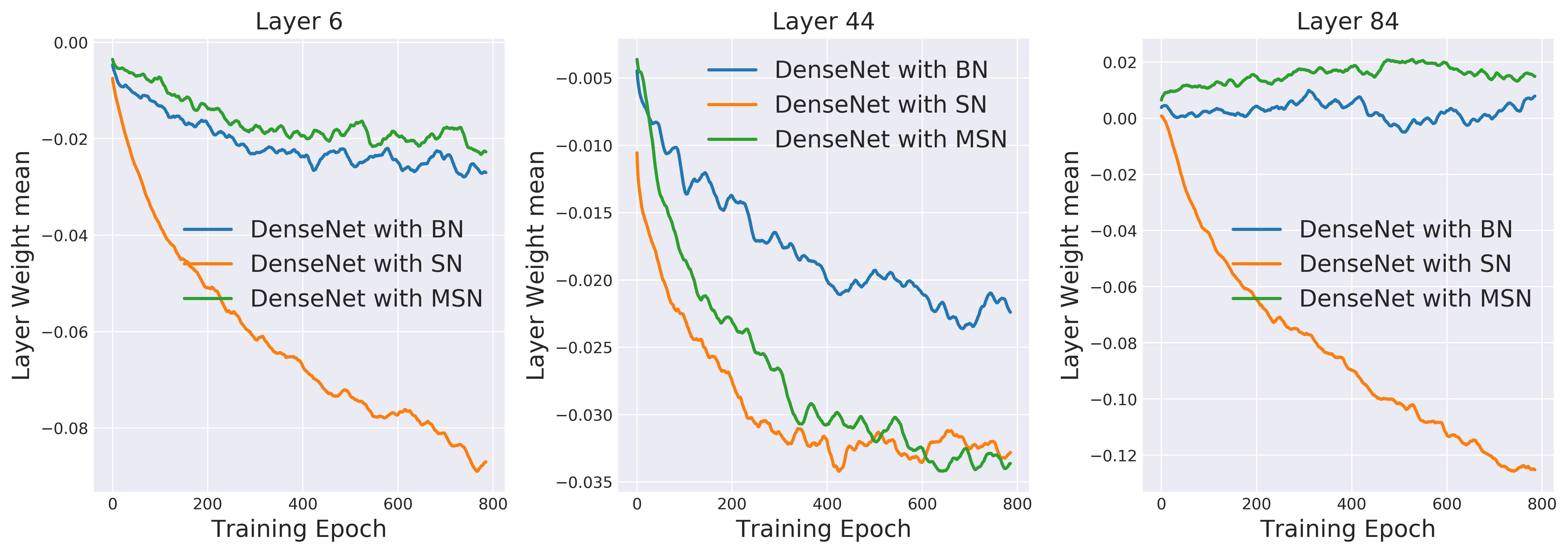}
    \caption{Mean drift correction by MSN during training DenseNet-BC. Our proposed MSN method reduces the gradual drift of the mean during training, which we hypothesize as the major cause for the reduced performance of SN networks compared to BN networks.}
    \label{fig:layer_mean_drift_cifar10}
\end{figure}

The empirical motivation for our proposed Mean Spectral 
Normalization is the reduced performance of SN for small and medium sized networks. Through our experimentation, we observed that the reason to be the gradual uncontrolled drift of the layer mean during training (Refer to Fig.  \ref{fig:layer_mean_drift_cifar10}). We hypothesize that the mean drift is directly related to the internal covariate shift, where the distribution of layer activations change during training. This can be clearly observed from Fig. \ref{fig:layer_mean_cifar10}, where the shift in some selected layers are shown. It is evident that the spectral norm sufficiently restricts the variance of the distribution of activations, however causes their mean to drift during training. Moreover, the mean-drift is also observable in batch normalized networks, but the drift is controlled by the bias $\beta_k$ during training. Therefore, the rapid and uncontrolled drift of the activation-distribution mean is the foremost cause for diminished performance of spectral normalized networks. We resolve this mean-drift by proposing a modification to the original SN, called as mean spectral normalization.

\subsection{Mean Spectral Normalization}
We explore the idea of combining SN with a part of BN, which we call Mean Spectral Normalization (MSN). In this method, we perform the spectral normalization on the weights and then subtract the minibatch means from the activations like with BN, as
\begin{align}\label{eq:MSN}
    \mathbf{h}_k &=  \frac{\mathbf{W}}{\sigma(\mathbf{W})}\mathbf{g}_{k-1}\\
    \tilde{\mathbf{h}}_k &= \mathbf{h}_k - \mathbb{E}[\mathbf{h}_k] + \mathbf{m}
\end{align}
where $\mathbf{h}_k$ is the preactivation for the given layer and $\mathbf{m}$ is the external bias learned during training. The activation $\mathbf{g}_k$ is then given by passing through the activation function $f$ as $\mathbf{g}_k = f(\tilde{\mathbf{h}}_k)$. By subtracting the mean, we create a normalization method that restricts the variance as well as the mean of the activation distribution, thereby resolving the problem of mean-drift. Moreover, the mean correction introduces only a small computational overhead compared to the full BN.

During training, the running average of the minibatch mean $\mathbb{E}[\cdot]$ is stored to be used for validation data.
The spectral norm of the weight matrix can be efficiently computed, with negligible overhead, using the power iteration in practice (as pointed out in \cite{miyato2018spectral}). During stochastic gradient descent, because the weights change slowly during each update, a single power iteration on the latest version of the initial vectors is sufficient for each training iteration; making MSN computationally more efficient than BN. By recentering the pre-activations, the dependency on the inputs of the neurons $\mathbf{x}$ on the pre-activations $\mathbf{h}_{l}$ is completely detached. This method of decoupling the norm of the pre-activations from the input vectors have shown to improve the rate of convergence \cite{salimans2016weight}.

We distinguish our MSN from the \textit{weight normalization with  mean-only batch norm} \cite{salimans2016weight} from the fact that unlike weight normalization, spectral norm does not reduce the rank of the weight matrix and therefore can leverage upon a wider range of features to improve the performance. Weight normalization, on the other hand, regularizes the network by forcing the network to produce weight matrices that lie (approximately) in low dimensional vector spaces, compromising the feature dimensions. Besides, by dividing by the Frobenius norm of the weights, weight normalization enforces a stronger restriction on the layer weights, often causing over-fitting. This was empirically shown in \cite{gitman2017comparison} where even other methods like dropout and weight decay failed to improve the generalization of the weight normalized network. Our proposed MSN, however, has a stronger regularizing effect than weight normalization as it restricts the layer weights in their gradient space, effectively regulating their learnability.

\subsection{Gradients of MSN}
The gradient of MSN can be computed as follows. Consider the gradient of the layer weight $\hat{\mathbf{W}}$ after SN, {\it w.r.t.} $\mathbf{W}_{ij}$
\begin{align}
    \frac{\partial \hat{\mathbf{W}}}{\partial \mathbf{W}_{ij}} &= \frac{1}{\sigma(\mathbf{W})} \bigg [ \mathbf{I}_{ij} - \frac{1}{\sigma(\mathbf{W})}\frac{\partial \sigma(\mathbf{W})}{\partial \mathbf{W}_{ij}}\mathbf{W} \bigg ]\\
    &= \frac{1}{\sigma(\mathbf{W})} \bigg [ \mathbf{I}_{ij} -[\mathbf{u}_1 \mathbf{v}^T_1]_{ij}\hat{\mathbf{W}} \bigg ]
\end{align}
where $\mathbf{I}_{ij}$ is the matrix that has $1$ in its $(i,j)^{th}$ entry and zero elsewhere; $\mathbf{u}_1$ and $\mathbf{v}_1$ are the left and right singular vectors of $\mathbf{W}$ respectively. Note that the first column of the left and right singular matrices of $\mathbf{W}$ correspond to the largest singular value of $\mathbf{W}$. Therefore, the gradient with respect to the largest singular value at a given element $\mathbf{W}_{ij}$ is the $(i,j)^{th}$ entry in the left and right singular vectors of the largest singular value.

Now, the gradient of the loss $\mathcal{L}$ with respect to MSN pre-activation $\mathbf{h}_k$ after the mean subtraction, can be found in a straightforward manner.
\begin{align}
    \nabla_{\mathbf{h}_k} \mathcal{L} = \nabla_{\tilde{\mathbf{h}}_k} \mathcal{L} - \mathbb{E} \big [ \nabla_{\tilde{\mathbf{h}}_k} \mathcal{L}\big ]
\end{align}
The recentering of the pre-activations has a much lower computational overhead compared to the classical BN where the second order batch statistics are required.

\begin{figure*}[ht]
    \centering
    \includegraphics[width = 0.8\textwidth, height=5.0cm]{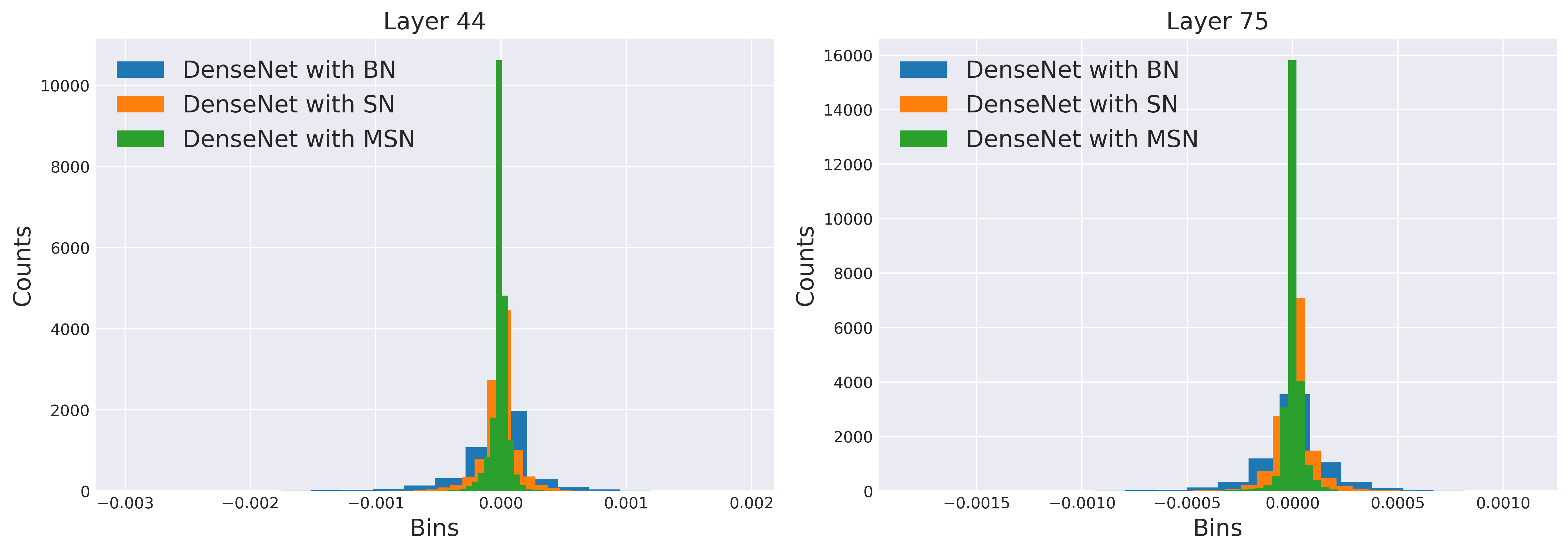}
    \caption{Gradient Histograms of layers 44 and 75 (chosen randomly) of DenseNet-BC with BN, SN and MSN respectively. For the spectral normalized network, the gradients are more concentrated around the mean compared to the batch normalized network. MSN clearly improves in restricting the gradients around the zero mean compared to BN or MSN.}
    \label{fig:gradient_hist_cifar10}
\end{figure*}

\begin{figure*}
    \centering
    \includegraphics[width = 0.9\textwidth, height=5.0cm]{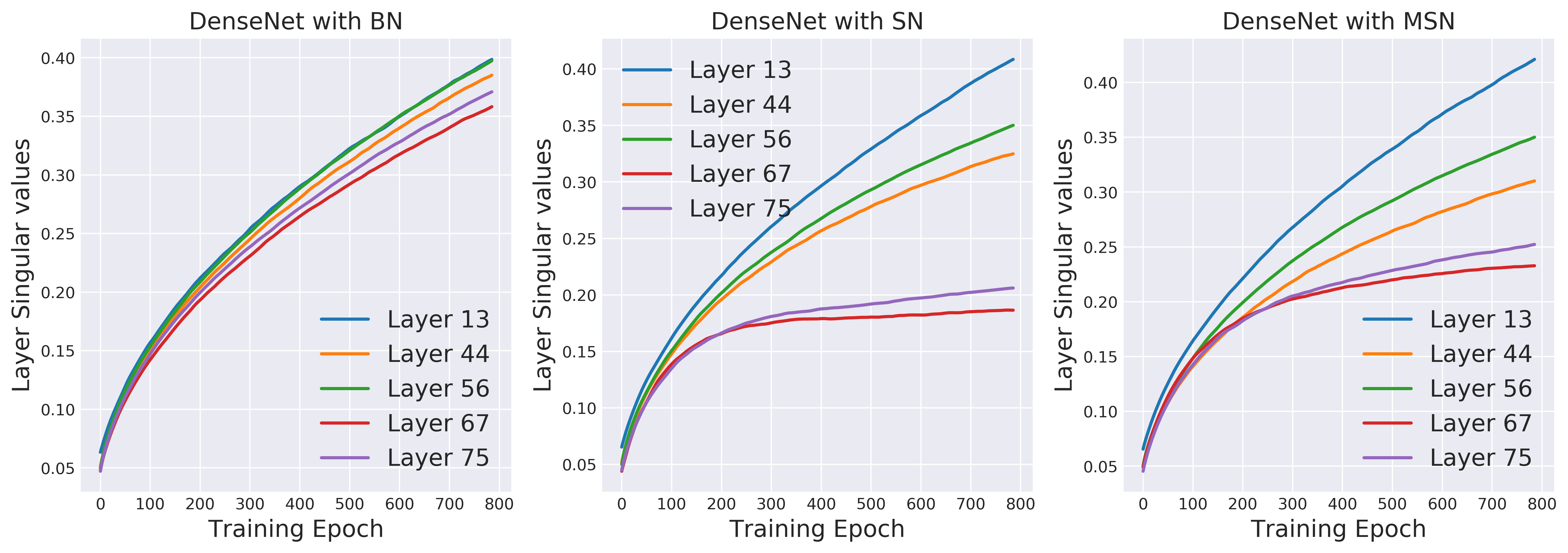}
    \caption{Mean layer-singular values during training for DenseNet-BC.  }
    \label{fig:layer_singular_cifar10}
\end{figure*}

\section{Experiments}\label{sec:exp}
\subsection{Experimental Setup}
To investigate the training dynamics of various normalization techniques discussed thus far, we use a set three different convolutions neural networks - 3-layer CNN (without pooling and dropout), VGG-7, and 100-layer DenseNet BC \cite{huang2017densely} architectures. These networks were trained on the standard MNIST\footnote{\href{http://yann.lecun.com/exdb/mnist/}{http://yann.lecun.com/exdb/mnist/}} , SVHN\footnote{\href{http://ufldl.stanford.edu/housenumbers/}{http://ufldl.stanford.edu/housenumbers/}} and CIFAR10\footnote{\href{ https://www.cs.toronto.edu/~kriz/cifar.html}{https://www.cs.toronto.edu/~kriz/cifar.html}} datasets, respectively. In this work, our core focus is on improving the networks for image-based tasks, as the application of DNNs in automation are predominantly image-based. We train these networks with Adam optimizer and set initial learning rates from $\{0.1, 0.001, 0.0001\}$. We train these models with sufficiently long epochs such that learning plateaus. We always report the best results among those learning rates. The code for all the experiments, plots and trained models are given in the following GitHub repository  $-$ \href{https://github.com/AntixK/mean-spectral-norm}{\color{RoyalBlue}https://github.com/AntixK/mean-spectral-norm}. Moreover, the choice of the networks was motivated from their widespread applications in real-world object recognition and segmentation.

\subsection{Training Dynamics and SN techniques}
In this section, we discuss our empirical observations of SN during training and the performance of our proposed normalization technique. Firstly, we present the performance comparison of BN, SN and MSN networks for all the three models in Fig. \ref{fig:test_acc_cifar10}, to illustrate the effectiveness of our proposed MSN weight reparametrization. We observe that MSN greatly improves upon SN for small and medium sized networks (3-layer CNN and VGG-7, respectively) and provides a comparable performance to that of BN. Table \ref{tab:test_acc} provides a comparison of test accuracy of all the models.

\justify
\textbf{Inducing sparsity -} Sparsity in DNNs has usually been  connected to its robustness, with the reasoning that the network automatically determines the right subset of parameters required to capture the high-level information from the data. From Fig. \ref{fig:sparse_gradients_cifar10}, it is evident that the SN and MSN methods constantly improve the gradient sparsity of the network during training, while the gradient sparsity in the batch normalized network saturates around $20\%$. One of the advantages of such sparse gradients is that they are well suited for distributed training of large neural networks \cite{jiang2018linear}, as little gradient information has to be shared between the sub-networks. Such distributed training of networks provide exciting opportunities for distributed training of autonomous systems. 

\justify
\textbf{Mean Drift Correction -} As discussed before, the mean-drift is a consequence of the internal covariate shift, observed in all neural networks in general. From Fig. \ref{fig:layer_mean_cifar10} and Fig. \ref{fig:layer_mean_drift_cifar10}, it is clear that the BN and MSN methods control the drift of the mean compared to SN.  We also observe that the mean-drift is always in the negative region. Large negative mean for layer weights causes the gradients to be extremely small after the LeakyReLU activation, used in all our models. This, in addition to already sparsified gradient, effectively reduced the learning capacity of the networks with many \textit{dead neurons}. BN avoids this problem with recentering its layer weights using a learnable bias $\beta_k$ (Refer to Eq. \ref{eq:BN}). In MSN, we follow a similar approach, where this recentering (Refer to Eq. \ref{eq:MSN}) avoids the mean-drift, causing a balance between creating a robust sparse network but preventing too many dead neurons. Fig. \ref{fig:layer_mean_drift_cifar10} confirms our hypothesis and MSN correctly rectifies the mean-drift to match the performance with that of batch normalised networks.

\justify
\textbf{Lipschitzness of the network -} From its definition, SN controls the Lipschitz constant of the hidden layers of the neural network to be $1$. Empirically, we observe this in our spectral normalized models as shown in Fig. \ref{fig:gradient_hist_cifar10}. Neural networks with SN and MSN methods, concentrate the gradients around the mean with much smaller variance compared to batch normalized neural networks with the same learning rate. As noted in \cite{bjorck2018understanding}, neural nets with larger variance of gradients or the models with heavy-tailed gradient histograms ({\it e.g.}, unnormalized networks) lead to divergence rather than convergence, as the training progresses.

Another interesting consequence of such concentrated gradients is that the loss landscape of the network becomes smooth as it does not allow erratic or sharp gradient changes. This smoothing of the loss landscape is one of the prime reasons for BN and now our proposed MSN to work over a wider range of learning rates.

\justify
\textbf{Singular value regularization -}
Fig. \ref{fig:layer_singular_cifar10} shows the variation of mean layer singular values of different layers during training.
Specifically, during training, BN causes the singular values to increase monotonically. Furthermore, the average layer singular values of all layers are closely spaced, implying that all the weight matrices lie on the same vector subspace. SN, on the other hand, causes the mean layer singular values to taper more quickly, especially for higher layers, where the the activations are affected by previous weight matrices. We reason that in spectral normalized networks, the weights of each layer lie in different vector subspaces and therefore has lesser freedom in choosing the number of singular components. In MSN, the bias correction term $\mathbf{m}$, improves the average singular value by appropriate factor, learned during training. As a result, the divergence of the mean singular values during training is reduced, forcing the weights to lie in the same vector subspace.

\begin{table}[ht]
\centering
\caption{Test accuracy comparison}
\begin{tabular}{llrrr}
\hline
Dataset & Model  & BN & SN & MSN \\ \hline
MNIST  & 3-layer CNN  & $98.28$ & $89.55$ & $96.71$   \\
SVHN & VGG-7 & 88.56& 78.43 &90.86\\
CIFAR10 & DenseNet-BC & $92.27$ & $90.52$ & $91.65$    \\ \hline
\end{tabular}
\label{tab:test_acc}
\end{table}

\begin{table}[t]
\centering
\caption{Qualitative comparison of MSNGAN, SNGAN and WGAN-GP}
\begin{tabular}{lll}
\hline
Model  & Inception Score (IS)  & FID Score \\ \hline
Real Data (CIFAR10)  & $11.24 \pm 0.12$  & $7.8$    \\ \hline
WGAN-GP(With BN)  & $6.42 \pm 0.10$  & $41.3$    \\
SNGAN  & $7.42 \pm 0.08$  & $29.3$    \\
MSNGAN (ours) & $7.39 \pm 0.07$ & $29.8$    \\ \hline
\end{tabular}
\label{tab:wgan}
\end{table}

\begin{table}[t]
\centering
\caption{Comparison of number of trainable parameters.}
\begin{tabular}{p{2cm}|l|p{3cm}}
\hline
Model                                 & Normalization             & Number of Parameters \\ \hline
\multirow{3}{2cm}{WGAN (Discriminator)} & BN  & $2,938,689$ \\
                                         & SN & $2,935,873\; (-2816)$ \\
                                         & MSN & $2,937,282 \; (-1407)$ \\\hline
\multirow{3}{*}{3-layer CNN}      & BN  &  $103,562$ \\
                                  & SN  &  $103,406\; (-156)$ \\
                                  & MSN & $103,486\; (-76)$  \\ \hline
\multirow{3}{*}{VGG-7}     & BN   & $438,314$    \\
                           & SN   &  $437,168 \;(-1146)$     \\
                           & MSN  &  $437,744 \;(-570)$                    \\\hline
\multirow{3}{*}{DenseNet-BC}   & BN &    $1,009,730$    \\
                               & SN &  $1,007,608 \;(-2122)$ \\
                               & MSN  & $1,008,670 \;(-1060)$   \\ \hline                 
\end{tabular}
\label{tab:num_param}
\end{table}

\justify
\textbf{Fewer trainable parameters -}
Table \ref{tab:num_param} shows the comparison of the number of trainable parameters of various models with BN, SN and MSN normalization methods. The amount of reduction in the number of parameters is given within parentheses. Note that the SN does not introduce any additional parameters to the network. Therefore, spectral normalized models have fewer trainable parameters compared to batch normalized models, and thus are usually faster during training and more memory efficient. Albeit introducing bias correction parameters, MSN still has lesser number of parameters compared to BN. Additionally, this reduction in number of parameters results in faster training. During our experiments, SN models trained $27\%$ faster compared to BN models, and MSN models trained $16\%$ faster than BN model. This reduction in the number of parameters, coupled with highly sparsity, makes the MSN a highly desirable choice for embedded application of DNNs. 

\subsection{Comparison with SNGAN}
We evaluate our proposed MSN method against the original SN method on the Wasserstein Generative Adversarial Network(WGAN) model (called as SNGAN \cite{miyato2018spectral}), for the task of unsupervised image generation on the CIFAR-10 dataset. Furthermore, we also compare these spectral norm-based Lipschitz regularizers against the gradient penalty (WGAN-GP) regularization \cite{gulrajani2017improved} method. Also, we use BN for WGAN-GP following the original paper. In this section, we shall very briefly discuss the GAN objective function and Lipschitz regularization.

The SN scheme was initially proposed for improving the training of Wasserstein GANs. Generative adversarial networks  \cite{goodfellow2014generative} are a class of generative models with two dueling neural networks - namely the generator $G$ and the discriminator $D$. The discriminator $D$ is trained to differentiate between real $\mathbf{x} \sim q_{data}(\mathbf{x})$ and fake data, while $G$ is trained to generate fake data $\mathbf{z}$ that $D$ identifies as real. However, in the original GAN, the gradient of the \textit{optimal} discriminator $D$ with respect to its input can be unbounded, and therefore can lead to instability in training or \textit{modal collapse}. Addressing this problem, various methods \cite{arjovsky2017wasserstein}, \cite{gulrajani2017improved} have been proposed for penalizing the Lipschitz constant -essentially regularising the gradients- of the discriminator in the form of Wasserstein distance-based GAN losses. Note that the Wasserstein distance, in its dual form, asserts that the discriminator function must have a Lipschitz constant of $K$. Thus, employing the Wasserstein distance rather than the original Jensen-Shannon distance for the GAN loss, implicitly requires that the discriminator gradients must be bounded. The WGAN objective function $V(G,D)$ used in our experiments (except for  WGAN-GP) is given as follows
\begin{align}\label{eq:SNGAN}
\begin{aligned}
    V(G,D) &= \underset{\mathbf{x} \sim q_{data}(\mathbf{x})}{\mathbb{E}} \big [\log D(\mathbf{x}) \big ]\\
    &+ \underset{\mathbf{z} \sim p(\mathbf{z})}{\mathbb{E}} \big [\log(1-D(G(\mathbf{z}))) \big ]
\end{aligned}
\end{align}

For WGAN-GP, we have an additional gradient penalty term, following the original paper. Furthermore, we observe that the recentering of the  pre-activations $\mathbf{h}$ in MSN, does not alter the Lipschitz norm of the activations $\mathbf{g}$. Therefore MSN still regularizes the Lipschitz norm of the activations effected by SN. 

We employ the same DCGAN \cite{radford2015unsupervised} architecture for both generator and discriminator as described in \cite{miyato2018spectral}.
To evaluate the quality of the generated image samples, we use the standard \textit{inception score} (IS) \cite{heusel2017gans} and the \textit{Fr\'{e}chet inception distance} (FID) \cite{salimans2016improved}. In Table \ref{tab:wgan}, we show the inception scores (higher, the better) and FID\footnote{Code obtained from \href{https://github.com/mseitzer/pytorch-fid}{https://github.com/mseitzer/pytorch-fid}} (lower, the better) for the unsupervised image generation on various models, with optimal setting, on the CIFAR10 dataset. The report the average scores over $5$ runs, each with $2000$ sampled images. The scores for the real CIFAR10 data is given for a baseline comparison. We observe that MSN clearly improves upon the WGAN-GP and performs at par with the original SNGAN.

\section{Discussion \& Conclusion}
Albeit originally proposed to control the Lipschitz constant of WGANs, we believe SN is a generic method to reparameterize the weights, with a goal to build a standardized framework to employ DNNs for robot automation.  In this paper, we investigated a consequence of the internal covariate shift, called mean drift, in spectral normalized networks, which affects their performance compared to BN. Furthermore, we presented many experimental results to demonstrate the gradient sparsity and Lipschitzness induced by SN in small, medium and large DNNs. We then proposed a solution to resolve the mean drift, called mean spectral normalization(MSN), deriving ideas from both BN and SN. Through our experiments, we confirm that MSN clearly out-performs SN for supervised classification models for all depths of neural networks. Parallelly, Farnia et al., \cite{farnia2018generalizable} observe a similar result as ours with spectral normalized DNNs. In contrast to our analysis, they conclude that the naive algorithm used to compute the spectral norm (the power iteration as used in \cite{miyato2018spectral}) was inefficient in regularizing the actual spectral norm of the convolution layers. To correct this, they slightly loosen the spectral norm constrain to be $\sigma(\mathbf{W}) \leq \beta$, where $\beta$ is some fixed value. Besides having no such tunable parameter, in our work, we observe a deeper mean-drift effect restricting network's performance and rectify the effect with our MSN method. Furthermore, we also compare the qualitative results of our MSNGAN with that of the SNGAN for unsupervised image generation. In future, we wish to focus on evaluating the performance of our MSN on sequence modelling tasks and on real-time data from robots.







\bibliographystyle{unsrt}
\bibliography{references}

\end{document}